\documentclass[letterpaper]{article} 
\usepackage{aaai24}  
\usepackage{times}  
\usepackage{helvet}  
\usepackage{courier}  
\usepackage[hyphens]{url}  
\usepackage{graphicx} 
\usepackage{natbib}  
\usepackage{caption} 
\usepackage{algorithm}
\usepackage{algorithmic}

\usepackage{url}            
\usepackage{booktabs}       
\usepackage{amsfonts}       
\usepackage{nicefrac}       
\usepackage{microtype}      
\usepackage{xcolor}         

\usepackage{amsmath}
\usepackage{amssymb}
\usepackage{newfloat}
\usepackage{listings}
\DeclareCaptionStyle{ruled}{labelfont=normalfont,labelsep=colon,strut=off} 
\floatstyle{ruled}
\newfloat{listing}{tb}{lst}{}
\floatname{listing}{Listing}
\newcommand{\etal}{\textit{et al}.}

\title{Semi-supervised Active Learning for Video Action Detection}
\author{
    Ayush Singh\textsuperscript{\rm 1}, Aayush J Rana\textsuperscript{\rm 2}, Akash Kumar\textsuperscript{\rm 2}, Shruti Vyas\textsuperscript{\rm 2}, Yogesh Singh Rawat\textsuperscript{\rm 2}\\
}
\affiliations{
    \textsuperscript{\rm 1} IIT (ISM) Dhanbad\\
    \textsuperscript{\rm 2}University of Central Florida\\
    ayush.s.18je0204@cse.iitism.ac.in, \{aayushjungbahadur.rana, akash.kumar, shruti, yogesh\}@ucf.edu

}

\usepackage{bibentry}

\begin{document}

\maketitle
\begin{abstract}
   In this work, we focus on label efficient learning for video action detection. We develop a novel semi-supervised active learning approach which utilizes both labeled as well as unlabeled data along with informative sample selection for action detection. Video action detection requires spatio-temporal localization along with classification, which poses several challenges for both active learning (\textit{informative sample selection}) as well as semi-supervised learning (\textit{pseudo label generation}).
   First, we propose \textbf{\textit{NoiseAug}}, a simple augmentation strategy which effectively selects informative samples for video action detection. Next, we propose \textbf{\textit{fft-attention}}, a novel technique based on high-pass filtering which enables effective utilization of pseudo label for SSL in video action detection by emphasizing on relevant activity region within a video. We evaluate the proposed approach on three different benchmark datasets, UCF-101-24, JHMDB-21, and Youtube-VOS. First, we   demonstrate its effectiveness on video action detection where the proposed approach outperforms prior works in semi-supervised and weakly-supervised learning along with several baseline approaches in both UCF101-24 and JHMDB-21.
   Next, we also show its effectiveness on Youtube-VOS for video object segmentation demonstrating its \textit{generalization capability} for other dense prediction tasks in videos. The code and models is publicly available at:  \url{https://github.com/AKASH2907/semi-sup-active-learning}. 
\end{abstract}

\section{Introduction}

\urldef{\myurl}\url{https://github.com/AKASH2907/semi_sup_active_learning}

Video understanding is an essential task for security, automation, and robotics \cite{rizve2021gabriella} as video data enables information extraction for detection \cite{hou2017real,yang2019step}, recognition \cite{hara2018resnet3d, kumar2023largescale}, tracking \cite{vondrick2018tracking}, and scene understanding \cite{lei2018tvqa}. 
Video understanding in general requires a large amount of labeled data to train an effective model. Collecting such data for dense prediction tasks such as action detection is even more challenging as it requires spatio-temporal annotations on every frame of the video.
In this work, we focus on label efficient learning for video action detection.




Existing works on label efficient learning for video action detection have primarily focused on weakly supervised learning, semi-supervised learning, and active learning. 
Weakly supervised methods often underperform compared to supervised methods, often requiring externally trained object detector to address the detection aspect of action detection \cite{cheron2018flexible,weinzaepfel2016human}. 
Recently, semi-supervised learning (SSL) \cite{kumar2022end} and active learning methods \cite{rana2022are} have shown promising performance. However, they also have their own limitations. SSL relies on randomly selected sub-samples, which can result in non-informative sample selection and suboptimal models. On the other hand, active learning aims to address this issue by selecting only informative samples for training. Nevertheless, it suffers from a cold-start problem, which makes it challenging to train a good model with limited labels which are initially available.





We propose a unified approach for video action detection by bridging the gap between semi-supervised learning (SSL) and active learning. We address the challenges of the cold-start problem in active learning by using an SSL technique to train a reliable initial model. Similarly, we resolve the need for informative training sample for SSL using optimized selection via active learning. Our student-teacher-based SSL framework benefits from active learning's informative sample selection, offering the advantages of both SSL and active learning for improved video action detection.

Video action detection needs to perform both spatio-temporal localization and classification. Solving this task using limited labels pose two distinct challenges; 1) determining the informativeness of samples, and 2) generating high-quality pseudo-labels. To address the first challenge, we propose \textit{\textbf{NoiseAug}}, a simple and novel augmentation strategy designed to estimate sample informativeness in video action detection. 
Model-driven AL used in existing works for sample selection often perturbs the model via regularization \cite{gal2016dropout,heilbron2018annotate,aghdam2019active}, which limits the extent of perturbation since too much perturbation will affect the network negatively. Therefore, we propose data-driven AL and use varying degree of data augmentations while maintaining video integrity. 
By isolating the type of augmentation seen by the model during training and selection step, we can focus on the relevant regions and reduce bias from training samples which is a common problem in active learning with limited labeled set \cite{pardo2021baod,gal2016dropout,aghdam2019active}. 

Active learning enables cost-effective labeling by selecting informative samples and subsequently helps improve model performance, benefiting scenarios where data annotation is expensive or time-consuming.
However, active learning also suffers from cold start when the initial model is trained using very limited labeled data \cite{houlsby2014cold,prabhu2019sampling}.
This leads to the second challenge, generating high-quality pseudo labels for semi-supervised learning (SSL). To tackle this, we introduce \textit{\textbf{fft-attention}}, a novel technique based on high-pass filtering that emphasizes on activity regions and their edges within a video. Fft-attention improves the prediction of activity regions and enhances the quality of pseudo labels for SSL in video action detection.

In summary, we make the following contributions,
\begin{itemize}
\setlength\itemsep{-1pt}
    \item We propose a novel semi-supervised active learning framework for video action detection which provides label efficient solution. To the best of our knowledge, this is the first work focusing on this problem.
    \item We propose \textit{NoiseAug}, a novel noise based augmentation for video data perturbation which helps in informative sample selection.
    \item We propose \textit{fft-attention}, a novel high pass filter which helps in estimating action and non-action regions for effective pseudo label generation in semi-supervised learning. 
\end{itemize}
We evaluate the proposed approach on two different video action detection benchmarks and compare with several baselines, including existing semi-supervised and weakly-supervised approaches  %
outperforming all prior works. 
We also demonstrate its effectiveness on Youtube-VOS for video object segmentation showing the generalization capability to other dense prediction tasks in videos.

\section{Related Work}

\paragraph{Video action detection}
Video action detection is a complex and challenging task \cite{yang2019step,pan2021actor,actor_centric,li2020actions, modi2022video}, where the goal is to perform spatio-temporal action detection in a given video. Most prior works use fully-supervised approach where all the samples are annotated spatio-temporally. The recent works have rapidly improved performance due to improved networks \cite{hara2018resnet3d,szegedy2017inceptionv4,he2016deep} and increased data availability \cite{soomro2012ucf101,jhuang2013towardsjhmdb}. However, getting large dataset with spatio-temporal annotation is costly. Weakly-supervised learning is an alternative which uses partially annotated data over the entire dataset to train action detection models \cite{mettes2017localizing,mettes2019pointly,cheron2018flexible,escorcia2020guess,arnab2020uncertainty,zhang2019glnet}. These methods rely on external pre-trained object detector \cite{frcnn} and often fall behind significantly on performance compared to the fully-supervised methods.

\paragraph{Semi-supervised learning}
Semi-supervised learning utilizes both labeled and unlabelled samples for training \cite{fixmatch,mixmatch,remixmatch,tarvainen2017mean,oliver2018realistic,miyato2018virtual,survey_semi,schiappa2022self}, generally using regularization \cite{const1,tarvainen2017mean,const3,const4} or pseudo-labeling \cite{pseudo_obj1, lee2013pseudo, rizve2021defense} methods for classification \cite{mixmatch,remixmatch,rizve2021defense, Dave_2023_CVPR} and detection \cite{kumar2022end,rosenberg2005semi}. For video action detection, using pseudo-labeling approach for semi-supervised learning becomes costly and difficult with limited labels \cite{zhang2022semi,schiappa2022self}. The pseudo-labeling approach also assumes that a pre-trained object detector or region proposal is available \cite{ren2020ufo,zhang2022semi}. 
A better option is to use consistency regularization which relies on the model itself to moderate the learning \cite{mixmatch,kumar2022end,const3,tarvainen2017mean,co_ssd}, generally using perturbations in input or model. We use a combination of consistency regularization via strong and weak augmentation of labeled and unlabeled samples using mean-teacher setup \cite{tarvainen2017mean}. One of the challenges in consistency based SSL for video action detection is having too much noise from background regions of a video, as seen in \cite{kumar2022end}. To this end, we focus on consistency of relevant action regions while suppressing large backgrounds present in videos with our fft-attention based filter approach. 

\paragraph{Active learning}



Labeling a large video dataset for action detection task is expensive as a lot of frames must be annotated spatio-temporally for each video. Active learning \cite{pardo2021baod} enables selecting samples for annotation by estimating the usefulness of each sample to the underlying task. It is used to iteratively select a subset of data for annotation on various tasks as image classification \cite{wang2016cost}, image object detection \cite{aghdam2019active,pardo2021baod} and video temporal localization \cite{heilbron2018annotate} with only few studies done for video action detection \cite{rana2022are}. The sample selection in AL is done using uncertainty \cite{liu2019deepAL}, entropy \cite{aghdam2019active}, core-set selection \cite{sener2017active} or mutual-information \cite{kirsch2019batchbald}. While there have been some prior works that combine AL and SSL for object detection and segmentation task \cite{elezi2022not,rangnekar2023semantic}, we are the first to propose a unified SSL active learning framework for spatio-temporal video action detection to best of our knowledge. We use data perturbation via noise based augmentation to get the model's uncertainty, using that as an estimate of usefulness for each sample in our AL strategy.


\begin{figure*}[t!]
    \centering
    \includegraphics[width=0.98\linewidth]{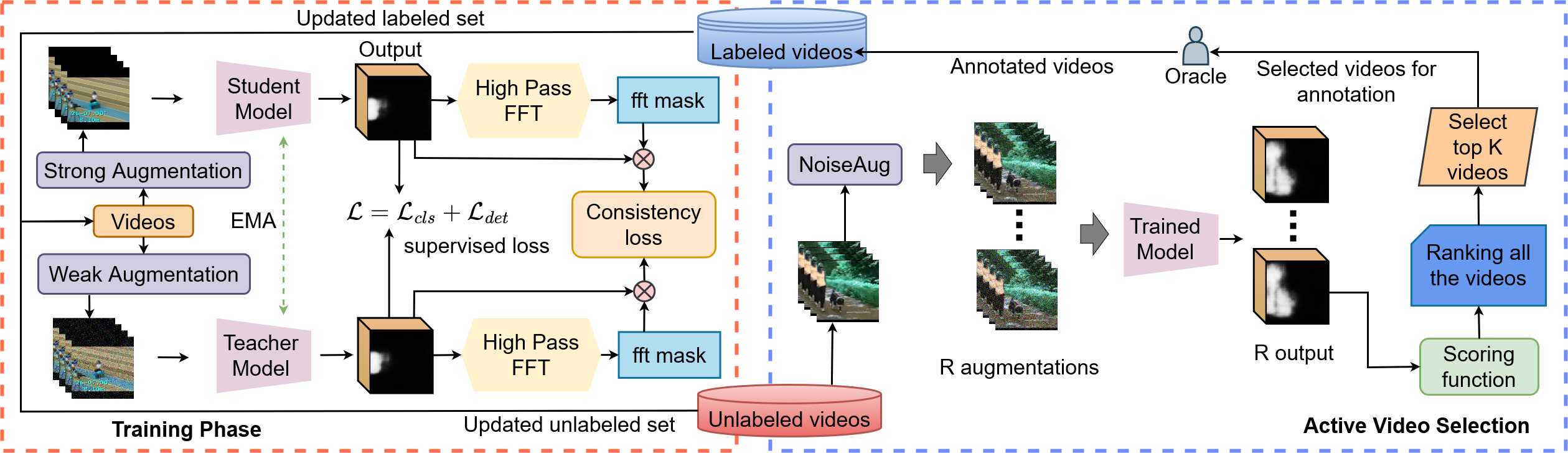}
    \caption{
    \textit{Overview of our proposed approach:} During the training phase, we take the labeled and unlabeled data at equal ratio to train the model together. We apply strong and weak augmentations to all input samples. All detection output is passed through our FFT filter to get a weight mask, which is used to compute the final consistency loss between teacher and student model output. During the active video selection phase, we take trained student model and pass $R$ variants with NoiseAug and score the sample. We select top K videos for further annotation and cycle back to the next training phase.}
    \label{fig:architecture}
\end{figure*}

\section{Proposed Method}

We introduce a semi-supervised active learning approach for video action detection, where active learning is employed to select samples during each training iteration, and SSL is utilized for model training. Our proposed NoiseAug strategy enhances sample selection in active learning. We use a student teacher based approach for SSL where we extend Mean Teacher \cite{tarvainen2017mean} for video action detection by incorporating fft-attention based filtering for effective pseudo-label training on unlabeled data. Mean Teacher approach relies on strong and weak augmentations to generate pseudo-labels that can be used to train augmented data. However, video data has a lot of unrelated background region that adds extra noise to this process. We propose fft-attention that helps focus on relevant activity region in videos and improves pseudo-label based training. An overview  of the  proposed approach is shown in Figure \ref{fig:architecture}.

\paragraph{Problem formulation:}
Given a dataset $\mathcal{D}$ with $N$ unlabeled videos $\mathcal{X}^U=\{x_1^U,...,x_N^U\}$ and $Q$ labeled videos $\mathcal{X}^L=\{x_1^L,...,x_Q^L\}$, we train an action detection model $M$ for $\theta$ weights using $\mathcal{D}=\{\mathcal{X}^L,\mathcal{X}^U\}$ using a semi-supervised approach. We assume that the initial labeled set is much smaller than the unlabeled set such that $Q<<N$. Once we have a trained model $M$, we use it for selecting more samples to annotate in the AL step. For each sample in $\mathcal{X}^U$, we prepare $\mathcal{V}$ variations by using proposed noise augmentation and use the model $M$ to estimate each sample's utility value. Once we do that for all $\mathcal{X}^U$, we rank them and select samples within budget for further annotation which makes a new labeled set $\mathcal{X}^L_2$ for training new model.
First, we describe the AL approach for sample selection from $\mathcal{X}^U$ for further annotation in next subsection.

\subsection{Active Learning for Sample Selection}
We use a trained model $M$ to get high utility samples from $\mathcal{X}^U$ for further annotation. This is an important step to increase labeled data under limited budget such that model training improves significantly over random sample selection approach. The model $M$ estimates prediction uncertainty of detection among all variations of a sample which in turn gives a sample level score on its usefulness. One of the key challenge to get this uncertainty is to avoid training bias of the network. Since the labeled training set $\mathcal{X}^L$ is often small initially ($Q<<N$), network can be easily over-fitted to have bias as well as become robust to the training augmentations. Our aim is to provide simple augmentation that is unique and not seen in the training time, which encourages the network to better estimate uncertainty of unlabeled samples. 

\paragraph{NoiseAug: } The goal is to find samples that maximize the model's performance for action detection. We follow prior AL works \cite{gal2016dropout,jain2009active} and use uncertainty as a measure for scoring and selecting samples. Prior works use Monte Carlo method \cite{gal2016dropout} or multi-layer output \cite{aghdam2019active} to compute per pixel informativeness (uncertainty, entropy). In contrast, we leverage 
on noise invariance to evaluate a sample's utility. We measure the variance in uncertainty of predictions from the model for different noise infused variants of the same sample. We generate multiple variations $\mathcal{V}_{\phi}$ of the same sample $v$ using noise augmentation given as,
\begin{equation}
\small
    \mathcal{V}_{\phi}^{i} = v^{[T\times H \times W \times C]} \odot \mathcal{N}^{i}(0,1)^{[T\times H \times W \times C]}
    \label{eq:noise_augmentation}
\end{equation}
where, $v$ is a sample with dimensions $[T\times H \times W \times C]$ and $\mathcal{N}^{i}(0,1)$ is a Gaussian distribution of the same dimension used as noise. We use their Hadamard product to get the final augmented variation $\mathcal{V}_{\phi}^{i}$. We repeat this process $R$ times to get a set of noise augmented variations $\mathcal{V}_{\phi}=\{\mathcal{V}_{\phi}^1,\mathcal{V}_{\phi}^2,...,\mathcal{V}_{\phi}^R\}$. 

\paragraph{Sample selection: } We use NoiseAug to measure uncertainty using model's confidence in each of the noise variants. We utilize the temporal aspect of a video for comparing pixel level uncertainty. We average the pixel values for neighboring frames as another form of regularization for uncertainty measure. We define the temporal average function as,
\begin{equation}
\small
    Avg(f_i[p]) = \frac{1}{T}\sum_{t=i-T/2}^{i+T/2} M(f_t[p];\theta)
    \label{eq:average_function}
\end{equation}
where, for a pixel $p$ in a given frame $f_i$ at the $i^{th}$ temporal location, we average the prediction of model $M$ with $\theta$ weights for same pixel location $p$ over neighboring $T$ frames. Then we compute the uncertainty score for sample $v$ as,
\begin{equation}
\small
    \mathcal{U}(v) = \sum_{i=1}^F \sum_{p=(1,1)}^{P_{[x,y]}} -log(Avg(f_i[p]))
    \label{eq:uncertainty_sample}
\end{equation}
where, $F$ is total frames with $[x,y]$ size, $M$ is the model with $\theta$ weights that gives prediction for the pixel $p$ in $i^{th}$ frame.

The sample's informativeness as a whole is a reflection of how consistent the model is to all of its noise variants. 
Ideally, the model $M$ is trained to be noise invariant. Any sample with high variance in uncertainty indicates that the network is not doing well for that sample when noise is introduced.
Thus, we use Equation \ref{eq:uncertainty_sample} on each of the $R$ variations to get the variance in uncertainty given as,
\begin{equation}
\small
    S = Var(\mathcal{U}[\mathcal{V}_{\phi}^1), \mathcal{U}(\mathcal{V}_{\phi}^2), ..., \mathcal{U}(\mathcal{V}_{\phi}^R)]  
    \label{eq:variance_uncertainty}
\end{equation}
where, $S$ is the informativeness score, $Var()$ gives the variance for uncertainty of all augmented variants $\mathcal{V}$ of sample $v$. For each AL round, we pick top $\textbf{K}$ samples for annotation such that our labeled videos becomes $\mathcal{X}^L_2=\{x^L_1,...,x^L_{Q+\textbf{K}}\}$ and our unlabeled set becomes $\mathcal{X}^U_2=\{x^U_1,...,x^U_{N-\textbf{K}}\}$. We use this new set of data $\mathcal{D}_2=\{\mathcal{X}^L_2,\mathcal{X}^U_2\}$ to train a new action detection model in the next round and continue this until we exhaust our total annotation budget.

\subsection{Semi-Supervised Learning}{\label{semi-supervised}}

To leverage the entire training dataset $\mathcal{D}=\{\mathcal{X}^L,\mathcal{X}^U\}$ with both labeled set $\mathcal{X}^L$ and unlabeled set $\mathcal{X}^U$, we use SSL approach that uses mean-teacher based regularization to train using the unlabeled data. Mean-teacher trains unlabeled data using pseudo-labels predicted from a teacher model on an augmented variation of the data. We use supervised loss on weak and strong augmented variations for the labeled data. 
For the unlabeled data, we use the prediction from teacher network $M_t$ to generate pseudo-labels that can be used to train the student network $M_s$ using supervised loss. Along with that, we also follow prior work \cite{kumar2022end} to use mean squared error based consistency loss for training $M_s$. For a given video $v$ with $F$ frames, we apply different degrees of augmentation following mean-teacher setup to obtain $v'$. The consistency loss is then computed as,

\begin{equation}
\small
    \mathcal{L}_{cons}^{MSE} = \frac{1}{F} \sum_{i=1}^{F} \frac{1}{x \cdot y} \sum_{p=(1,1)}^{P_{[x,y]}} \| M_s(f_p^{i};\theta_s) - M_t(f_p^{i'};\theta_t)\|^{2_2}
    \label{eq:consistency_loss}
\end{equation}

where, $M_t$ and $M_s$ are teacher and student models with $\theta_t$ and $\theta_s$ weights respectively that gives spatio-temporal detection for $i^{th}$ frame. We compute MSE value for each pixel $p$ for frame $f^i$ of $[x,y]$ size.
This general form of MSE based consistency loss gives equal weight for all $P$ pixels in the frame, which is non-ideal for spatio-temporal detection as we only want to focus on certain action regions in each frame $f$. 

It is preferable to focus on relevant regions without using manually designed heuristics (pre-computed regions \cite{frcnn}). We want to reduce model uncertainty for specific areas with lower prediction quality, specifically the edges of an actor. To this end, we propose using a high pass filter that will reduce low frequency areas and give more focus on the high frequency area such as edges. Next, we define the high pass filter we use for selective focus.

\begin{table*}[t!]
\small
  
  \centering
  \begin{tabular}{c| c c | c c c c | c c c c}

    \toprule
    Method & \multicolumn{2}{c|}{Backbone}  & \multicolumn{4}{c|}{UCF101-24}  & \multicolumn{4}{c}{JHMDB-21}  \\
    && & Label & f-mAP & \multicolumn{2}{c|}{v-mAP} & Label & f-mAP & \multicolumn{2}{c}{v-mAP}\\
    
    \midrule
     & 2D & 3D & \% &   0.5  & 0.2 & 0.5 & \% & 0.5 & 0.2 & 0.5 \\
    \midrule
    \textbf{Fully-Supervised} &&&&&&&& \\
    \midrule 
    Kalogeitan \etal \cite{kalogeiton_2017} & \checkmark& & & 69.5 & 76.5 & 49.2 & &  65.7 & 74.2 & 73.7  \\
    Song \etal \cite{Song_2019_CVPR}$^{\dagger}$ & \checkmark& & & 72.1 & 77.5 & 52.9 & &  65.5 & 74.1 & 73.4  \\
    Li \etal \cite{li2020actions}  & \checkmark& &  & 78.0 & 82.8 & 53.8 & & 70.8 & 77.3 & 70.2\\
    Gu \etal \cite{ava}$^{\dagger}$ & & \checkmark& & 76.3 & - & 59.9& & 73.3 & - & 78.6 \\
    Duarte \etal \cite{duarte2018videocapsulenet} & & \checkmark & & 78.6 & \underline{97.1} & \underline{80.3} & & 64.6 & \underline{95.1} & -   \\
    Pan \etal \cite{pan2021actor} & & \checkmark & & \underline{84.3} & -& -& & - & - & - \\
     Zhao \textit{et al.} \cite{Zhao_2022_CVPR} & & \checkmark & &83.2 & 83.3 & 58.4 &&  - & 87.4 & \underline{82.3}  \\
     Wu \textit{et al.} \cite{Wu2023STMixerAO} &  & \checkmark & &83.7 & - & - && \underline{86.7} & - & -\\
    \midrule
    \textbf{Weakly-Supervised} &&&&&&&& \\
    \midrule
    
    Mettes \etal \cite{mettes2017localizing} & \checkmark & & &- &37.4&-  & &- &-&  -\\
    Mettes and Snoek \cite{mettes2019pointly} & \checkmark& & &- &41.8&- & & - &-&  -\\
    Cheron \etal \cite{cheron2018flexible} & & \checkmark& & - & 43.9  & 17.7 & &- & - & -  \\
    Escorcia \etal \cite{escorcia2020guess} & & \checkmark & & 45.8 & 19.3 & - & & - & - & - \\
    Arnab \etal \cite{arnab2020uncertainty} & & \checkmark& & - & 61.7 & 35.0 & & - & - & -  \\
    Zhang \etal \cite{zhang2019glnet} & & \checkmark & & 30.4 & 45.5 & 17.3& & 65.9 & 77.3 & 50.8 \\
    
    \midrule
    \textbf{Semi-Supervised} &&&&&&&& \\
    \midrule
    MixMatch \cite{mixmatch} && \checkmark & 20\% & 20.2 & 60.2 & 13.8 & 30\% & 7.5 & 46.2 &  5.8 \\
    Psuedo-label \cite{lee2013pseudo} && \checkmark & 20\% & 64.9 & 93.0 & 65.6 & 30\% &  57.4 & 90.1 & 57.4\\
    Co-SSD (CC)\cite{co_ssd} & & \checkmark& 20\% & 65.3 & 93.7 & 67.5 & 30\% &  60.7& 94.3& 58.5 \\
    PI-consistency \cite{kumar2022end} & & \checkmark& 20\% & 69.9 & 95.7 & 72.1 & 30\% &  64.4 &95.4 & 63.5  \\
    \hline
    Ours (M-T SSL)              & & \checkmark  & 20\% & {69.8} & {94.9} & {72.2} & 30\% &  {68.5} & {98.4} & {68.0}   \\
    Ours (M-T SSL + AL)         & & \checkmark  & 20\% & \textbf{72.0} &\textbf{96.3}  & \textbf{74.5} & 30\% &  \textbf{70.7} &\textbf{98.8} & \textbf{71.7}   \\
    \midrule
    Supervised baseline   & & \checkmark  & 20\% & 59.8   &  91.6     &  59.2  & 30\% &    59.4  &  96.5  &  60.4  \\
    \bottomrule

  \end{tabular}
    \caption{\textit{Comparison with existing works}: We compare 
  with existing supervised and weakly supervised works along with the semi-supervised baselines on UCF101- 24 and JHMDB-21. $\dagger$ denotes method using Optical flow.
}
  \label{tab:ssl_main}
\end{table*}

\paragraph{FFT based high pass filter}
In order to separate the low and high frequency areas, we apply a FFT based high pass filter. We are trying to focus more on the edges of the predicted detection regions while suppressing other areas. We assume that the high frequency areas (edges) are harder for the network to learn due to quick changes at the edges in a video. Thus, we identify such regions and give them higher weight than the easier regions during training to increase model's consistency. A FFT high pass filter finds the edges and attenuates lower frequency, keeping the non-edge regions with lower weight. We define the FFT high pass filter function as, 
\begin{equation}
\small
    HPF(f) = FFT(M(f;\theta))
    \label{eq:fft_function}
\end{equation}
where, $M$ is the model with weight $\theta$, $FFT()$ is the FFT function that gives the filtered output for a frame $f$. For a given video $v$ and its augmented variant $v'$ with $F$ frames, the per frame consistency using the $FFT$ filter is,
\begin{equation}
\small
    FC(f,f',W) = \frac{1}{x.y}\sum_{p=(1,1)}^{P_{[x,y]}} \|M_s(f_p;\theta_s) - M_t(f_p';\theta_t)\|^2_2 \cdot W_p
    \label{eq:frame_hpf_consistency}
\end{equation}
where, $FC(f,f',W)$ is the frame-wise consistency function that takes frame $f$, $f'$ and weight $W$, all of $[x,y]$ size. We modify the MSE computation from Equation \ref{eq:consistency_loss} to use pixel-wise weight $W$ on the computed MSE value of pixel $p$. Then we redefine the overall consistency loss as,
\begin{equation}
\small
    \mathcal{L}_{cons}^{HPF} = \frac{1}{F} \sum_{i=1}^{F} FC(f^{i}, f^{i'}, HPF(f^{i}))
    \label{eq:hpf_consistency_loss_1}
\end{equation}
\begin{equation}
\small
    \mathcal{L}_{cons}^{HPF'} = \frac{1}{F} \sum_{i=1}^{F} FC(f^{i}, f^{i'}, HPF(f^{i'}))
    \label{eq:hpf_consistency_loss_2}
\end{equation}
where, we compute the per-frame consistency with $HPF$ as weight from both $f$ and $f'$ frames of $v$ and $v'$ videos.

\paragraph{Temporal consistency}
We use the temporal information of subsequent frames to improve the consistency loss for training. While the $HPF$ based consistency computes spatial consistency for each frame, it does not use temporal consistency information. To enforce temporal consistency, we use temporal average function from Equation \ref{eq:average_function}, which changes the model's output for Equation \ref{eq:fft_function} and \ref{eq:frame_hpf_consistency} from $M(f;\theta) \rightarrow Avg(f)$.

\subsection{Overall Training Objective}

We train the model $M_s$ with $\mathcal{D}=\{\mathcal{X}^L,\mathcal{X}^U\}$ which consists of both labeled and unlabeled data. We use supervised loss on labeled data for classification $\mathcal{L}_{cls}$ and detection $\mathcal{L}_{det}$. For the unlabeled data, we use $M_t$ to get pseudo-label for training $M_s$ in a supervised fashion. For unsupervised $M_s$ training, we use the proposed FFT high pass filter based consistency loss. Our training objective is given as,
\begin{equation}
\small
    \mathcal{L}^{overall}_{cons} = \lambda_1\mathcal{L}^{HPF}_{cons} + \lambda_2\mathcal{L}^{HPF}_{cons'}
    \label{eq:overall_const_loss}
\end{equation}
\begin{equation}
\small
    \mathcal{L} = \mathcal{L}_{cls} + \mathcal{L}_{det} + \mathcal{L}^{overall}_{cons}
    \label{eq:overall_loss}
\end{equation}
where, $\lambda_1$ and $\lambda_1$ are loss weights given to the consistency loss which varies for unlabeled samples following prior SSL works \cite{fixmatch,kumar2022end}.

\section{Experiments}

\paragraph{Datasets }
We conduct our experiments on three video datasets, UCF101-24 \cite{soomro2012ucf101} and JHMDB-21 \cite{jhuang2013towardsjhmdb}. UCF101-24 consists of 24 classes with a total of 3207 untrimmmed videos with bounding box annotations. JHMDB-21 dataset has 21 classes from a total of 928 videos with pixel-level annotations. Both UCF101-24 and JHMDB-21 are focused on action detection task. 
We further generalize our approach on YouTube-VOS dataset, a video object segmentation task, which has temporally sparse pixel-wise mask annotation for specific objects. It has 3471 videos for training with 65 object categories.



\begin{figure*}[t!]
    \centering
    \includegraphics[width=0.9\linewidth]{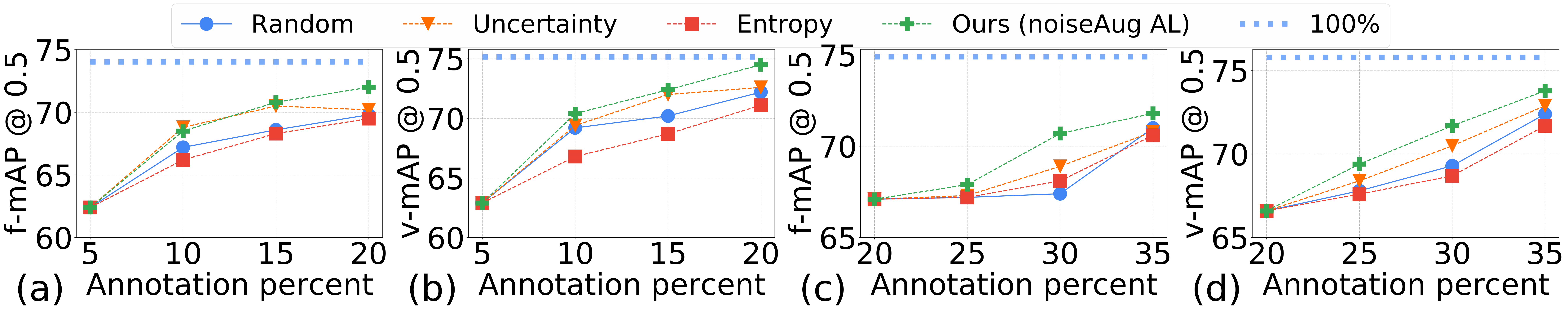}
    \caption{\textit{Analysis on selection criteria}: 
    We compare our proposed AL selection with other selection baselines using the same SSL training setup on UCF101-24 (a-b) and JHMDB-21 (c-d). 
    }
    \label{fig:baseline_al_ucf101_jhmdb}
\end{figure*}

\begin{figure*}[t!]
\centering
    \includegraphics[width=.09 \linewidth,height=.07 \linewidth]{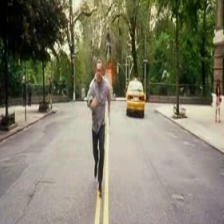} 
    \includegraphics[width=.09 \linewidth,height=.07 \linewidth]{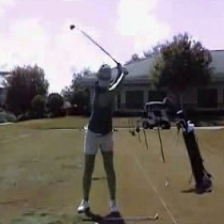} 
    \includegraphics[width=.09 \linewidth,height=.07 \linewidth]{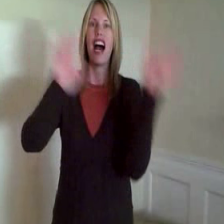} 
    \includegraphics[width=.09 \linewidth,height=.07 \linewidth]{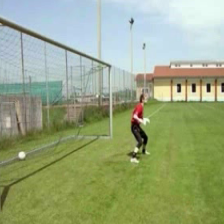}
    \includegraphics[width=.09 \linewidth,height=.07 \linewidth]{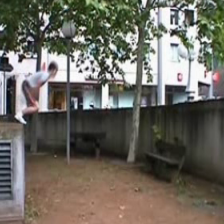} 
    \includegraphics[width=.09 \linewidth,height=.07 \linewidth]{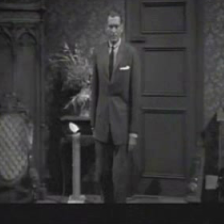}
    \includegraphics[width=.09 \linewidth,height=.07 \linewidth]{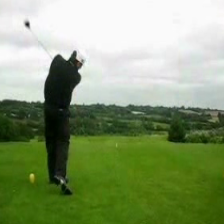}
    \includegraphics[width=.09 \linewidth,height=.07 \linewidth]{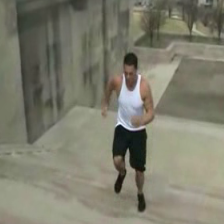}
    \includegraphics[width=0.09 \linewidth,height=0.07 \linewidth]{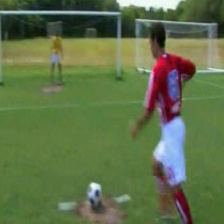}  
    \includegraphics[width=0.09 \linewidth,height=0.07 \linewidth]{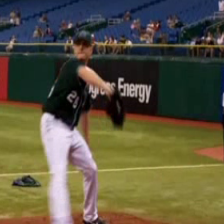}
    \\
    \includegraphics[width=.09 \linewidth,height=.07 \linewidth]{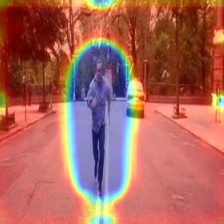}
    \includegraphics[width=.09 \linewidth,height=.07 \linewidth]{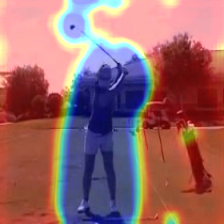}
    \includegraphics[width=.09 \linewidth,height=.07 \linewidth]{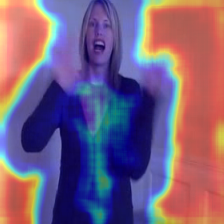}
    \includegraphics[width=.09 \linewidth,height=.07 \linewidth]{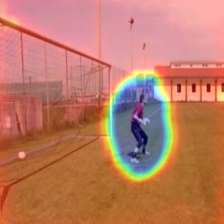}
    \includegraphics[width=.09 \linewidth,height=.07 \linewidth]{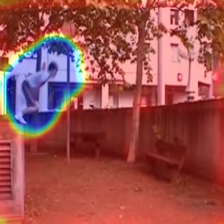}
    \includegraphics[width=.09 \linewidth,height=.07 \linewidth]{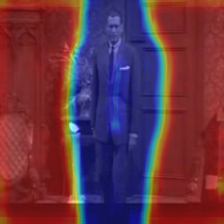}
    \includegraphics[width=.09 \linewidth,height=.07 \linewidth]{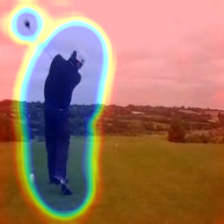}
    \includegraphics[width=.09 \linewidth,height=.07 \linewidth]{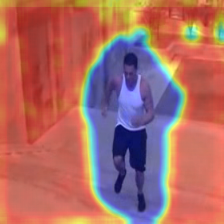}
    \includegraphics[width=0.09 \linewidth,height=0.07 \linewidth]{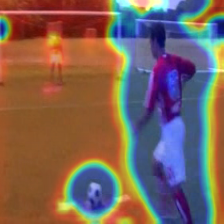} 
    \includegraphics[width=0.09 \linewidth,height=0.07 \linewidth]{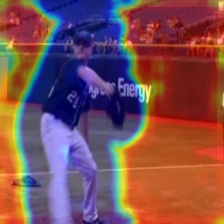} 
\caption{\textit{Qualitative analysis of FFT based high pass filter:} We show the input frames (first row) and corresponding weights (bottom row) using proposed FFT filter. The FFT method gives higher weight towards the edges of detected action regions while suppressing background. Red: low weight, blue: medium weight, green: high weight}.
\label{fig:heatmap}
\end{figure*}

\begin{table*}[t!]
  \small
  
  \centering
  \setlength\tabcolsep{3pt}
  \begin{tabular}{ ccc | cc|cc|cc||cc|cc|cc}
    \toprule
      & & & \multicolumn{6}{c||}{UCF101-24}  &  \multicolumn{6}{c}{JHMDB-21} \\
    \hline
     \multicolumn{3}{c|}{} & \multicolumn{2}{c|}{10\%} & \multicolumn{2}{c|}{15\%} & \multicolumn{2}{c||}{20\%} & \multicolumn{2}{c|}{20\%} & \multicolumn{2}{c|}{25\%} & \multicolumn{2}{c}{30\%} \\
    
      C  &  M-T  &  FFT  & f-mAP & v-mAP & f-mAP & v-mAP & f-mAP & v-mAP & f-mAP & v-mAP  & f-mAP & v-mAP & f-mAP & v-mAP  \\
      
    \midrule
    
    \checkmark  &             &              &  62.4  &  62.5    &  64.6  &  65.3  &  66.5  &  68.7  &  62.6	  &  59.3  &  63.1  &  62.9  &  63.4  &  64.2  \\
    & \checkmark  &              &  67.2  &  68.6	 &  68.4  &  69.5  &  69.2  &  71.9  &  61.5	  &  63.3  &  65.6  &  65.3	 &  66.0  &  67.4 \\
    & \checkmark  &  \checkmark  &  68.5  &  70.4 &  70.8  &  72.4  &  72.0  &  74.5  &  67.1	  &  66.6  &  65.2	&  67.8	 &  67.4  &  69.3  \\
      
    \bottomrule
    
  \end{tabular}
  \caption{\textit{Ablations}: We show effectiveness of different components used in SSL training. We evaluate the effect of consistency based SSL (C), mean-teacher (M-T) setup and proposed FFT filter during the training. We report f-mAP and v-mAP @ 0.5.
  }
  \label{tab:ab_ssl_combined}
\end{table*}

\paragraph{Evaluation metrics}
Following prior action detection works \cite{Peng_2016} we evaluate the f-mAP and v-mAP scores at different IoU thresholds for UCF101-24 and JHMDB-21. The f-mAP is computed from spatial IoU for each frame per class and averaged for all frames to get precision score at given IoU. Similarly, v-mAP is computed using spatio-temporal IoU for each video per class and averaged. For VOS task, we compute the average IoU and boundary similarity score following \cite{vosdataset}.

\paragraph{Implementation details:}
We use the PyTorch to build our models and train them on single 16GB GPU. For action detection, we use VideoCapsuleNet \cite{duarte2018videocapsulenet,kumar2022end,rana2022are}, with margin-loss for classification and BCE loss for detection. The network input is $8$ RGB frames of size $224 \times 224$. We use a batch size of $8$ for training with equal ratio of labeled and unlabeled sample per batch. We use the Adam optimizer \cite{kingma2014adam} with a learning rate of $1e-4$. We use EMA update at rate of 0.996. We train UCF101-24 for 80 epochs and JHMDB-51 for 50 epochs. \textbf{Hyperparameters:} We use a temporal block of $T=3$ frames for the temporal average function in Equation \ref{eq:average_function}. 
The loss weights for the consistency loss are  $\lambda_1=0.5$ and $\lambda_2=0.5$ in Equation \ref{eq:overall_const_loss} and $\lambda_3=[0.01\rightarrow0.1]$ increased over warmup range. Please refer to supplementary for more details. \textbf{Active Learning:} We take $R=8$ different noise added variations for each video $v$ to get the sample informativeness score $S$ in Equation \ref{eq:variance_uncertainty}. We select $5\%$ and $2\%$ new samples for UCF101-24 and $10\%$ for JHMDB-21 in each AL round. We take temporal average over $T=3$ frames in Equation \ref{eq:average_function}.

\subsection{Active Learning Baselines}

We compare our proposed AL approach with baselines on both UCF101-24 and JHMDB-21. We use random selection, MC uncertainty \cite{gal2016dropout}, MC entropy \cite{aghdam2019active} as selection baselines to compare with the proposed AL selection. All baselines use same backbone as ours for fair comparison, with results in Figure \ref{fig:baseline_al_ucf101_jhmdb}. 

\paragraph{UCF101-24}
We begin with 5\% labeled data and increment by 5\% in every AL cycle as shown in Figure \ref{fig:baseline_al_ucf101_jhmdb}(a-b). With more data, we notice that compared to baseline selection methods, our AL method is consistently performing better. We also notice a cold start problem for MC entropy as the model is not performing well for most samples in initial round of 10\%, using only model entropy leads to non-optimum sample selection in future rounds. Our AL approach uses noise augmentation to estimate the model uncertainty along with temporal averaging to utilize the temporal consistency expected in videos, which leads to better sample informativeness scores that is more reflective of the model's need.

\paragraph{JHMDB-21}
Due to the dataset being smaller with only 660 training videos and the detection task being harder with pixel-wise semantic segmentation, we initialize the training with 20\% labeled data ($198$ videos) for all methods. Each round increases labeled data by 5\% videos until we reach 35\% data. The quantitative results are shown in figure \ref{fig:baseline_al_ucf101_jhmdb}(c-d). Similar with UCF101-24, we see that our AL method consistently outperforms baseline selection methods.

\subsection{SSL Baselines}
We compare the effect of different SSL techniques to show why the proposed mean-teacher setup is optimum for video understanding task. We use mean-teacher SSL and compare results with consistency based SSL in Table \ref{tab:ab_ssl_combined}. We observe that mean-teacher based SSL outperforms consistency based SSL for all dataset, showing that the controlled weight update of teacher using EMA and pseudo-label from teacher using augmented data better regulates unlabeled training.

\subsection{Comparison With the State-of-the-Art}
We compare to prior works using fully, weakly and semi-supervised approach on UCF101-24 and JHMDB-21 in Table \ref{tab:ssl_main}. Compared to the weakly supervised methods, we perform significantly better for both dataset. The semi-supervised methods are closer in performance with our approach. We show that our method with only mean-teacher (M-T) SSL (no AL selection) performs better than existing SSL methods. The temporal consistency component and focusing on edges using FFT high pass filter enables our model to weight the relevant regions appropriately compared to background regions, which gives our method a competitive edge. Furthermore, when we use the proposed noise augmentation based AL to do sample selection and train using SSL, we perform better than prior weakly and semi-supervised methods with $+2.4\%$ v-mAP@0.5 for UCF101-24 and $+8.2\%$ v-mAP@0.5 for JHMDB-21 over prior best score. 
We also show the generalization on YouTube-VOS in Table \ref{tab:vos_experiments}.

\begin{table}[t!]
    \small
  \centering
  \begin{tabular}{ccccccc}
   \toprule
    Method & Data & Avg & $J_{S}$ & $J_{U}$ & $F_{S}$ & $F_{U}$ \\
    \midrule
    Random & 10\% & 10.1 & 11.6 & 10.1 & 9.6 & 9.2 \\
    PI-consistency & 10\% & 36.8 & 43.1 & 31.4 & 40.8 & 31.8 \\
    Ours   & 10\% & 39.3 & 46.1 & 33.7 & 43.9 & 33.5\\
    \hline
     Random & 20\% & 34.7 & 42.8 & 29.0 & 38.7 &28.3 \\
     Ours               & 20\% & 41.6 & 49.6 & 34.2 & 47.7 & 35.6 \\
    \bottomrule
  \end{tabular}
  \caption{\textit{Generalization capability}: Evaluation on Youtube-VOS dataset. We use same backbone following \cite{xu2018youtube} for our and random method. 10\% results for PI-consistency are reported from \cite{kumar2022end}.}
  \label{tab:vos_experiments}
\end{table}

\subsection{Ablations}

\begin{table}[t!]
  \small
  
  \centering
  \setlength\tabcolsep{3pt}
  \begin{tabular}{ cc | cc|cc|cc|cc}
    \toprule
     \multicolumn{2}{c|}{} & \multicolumn{2}{c|}{5\%} & \multicolumn{2}{c|}{10\%} & \multicolumn{2}{c|}{15\%} & \multicolumn{2}{c}{20\%} \\

    S/W  &  noiseAug  & \textit{f} & \textit{v} & \textit{f} & \textit{v} & \textit{f} & \textit{v} & \textit{f} & \textit{v}   \\
      
    \midrule
    
    \checkmark  &              &  62.4  &  62.9  &  61.9  &  62.3  &  63.6  &  63.4  &  64.0  &  64.2  \\
                &  \checkmark  &  62.4  &  62.9  &  68.5  &  70.4  &  70.8  &  72.4  &  72.0  &  74.5  \\
    \bottomrule
    
  \end{tabular}
  \caption{\textit{Effectiveness of NoiseAug}: Comparison of different augmentations used for AL selection for strong/weak augmentation from mean teacher SSL training and proposed noiseAug. [\textit{f}: f-mAP, \textit{v}: v-mAP @ 0.5]} 
  \label{tab:ab_al}
\end{table}

\paragraph{Effect of FFT}
We evaluate the usefulness of FFT filter as weights for putting more emphasis on regions around an action. We train the teacher and student model for action detection without FFT and compare with our baseline in Table \ref{tab:ab_ssl_combined}. We observe a drop in performance when the FFT filter is not used as a weight to compute the consistency loss, showing that the regions selected using FFT have more relevance to action detection. We also see how the FFT filter emphasises relevant regions during training in Figure \ref{fig:heatmap}.

\paragraph{Augmentations for AL consistency}
We use noise augmentation for our AL selection strategy, where we use \textbf{$R=8$} noise augmented variants of the video $v$ to compute uncertainty variance. To validate the effectiveness of this augmentation, we use the same strong/weak augmentation setup used for mean-teacher SSL training process (details in suppementary). As we observe in Table \ref{tab:ab_al}, using the proposed noise based augmentation provides useful information for sample selection compared to using strong/weak augmentation from SSL step. The network is already trained with strong/weak augmentation, making the predictions more robust for such augmentations in the AL selection step. Varying levels of noise-based augmentation enable the network to encounter different sample variations in AL selection from training, leading to improved uncertainty estimation for new samples.



\subsection{Discussion and Analysis}


\begin{figure}[t!]
    \centering
    \includegraphics[trim=0.8cm 0.5cm 0.5cm 0.7cm, width=0.48\linewidth]{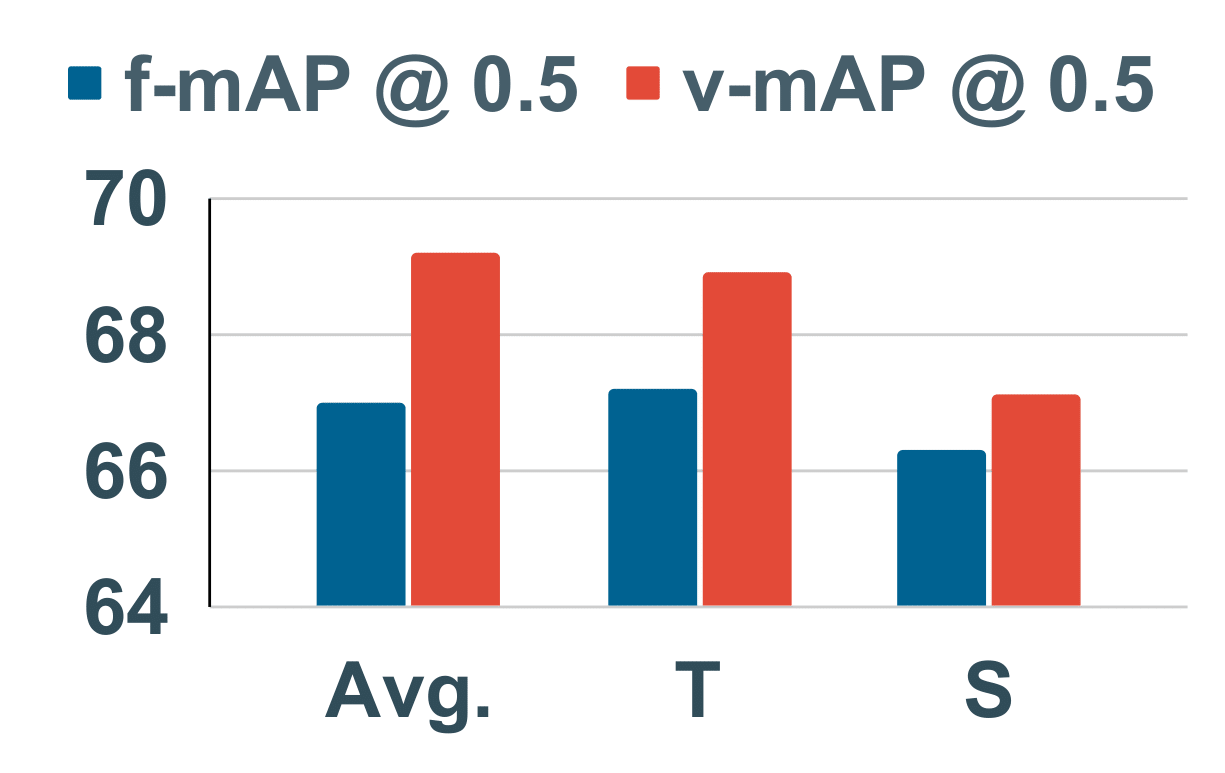}
    \includegraphics[trim=0.8cm 0.5cm 0.5cm 0.8cm, width=0.48\linewidth]{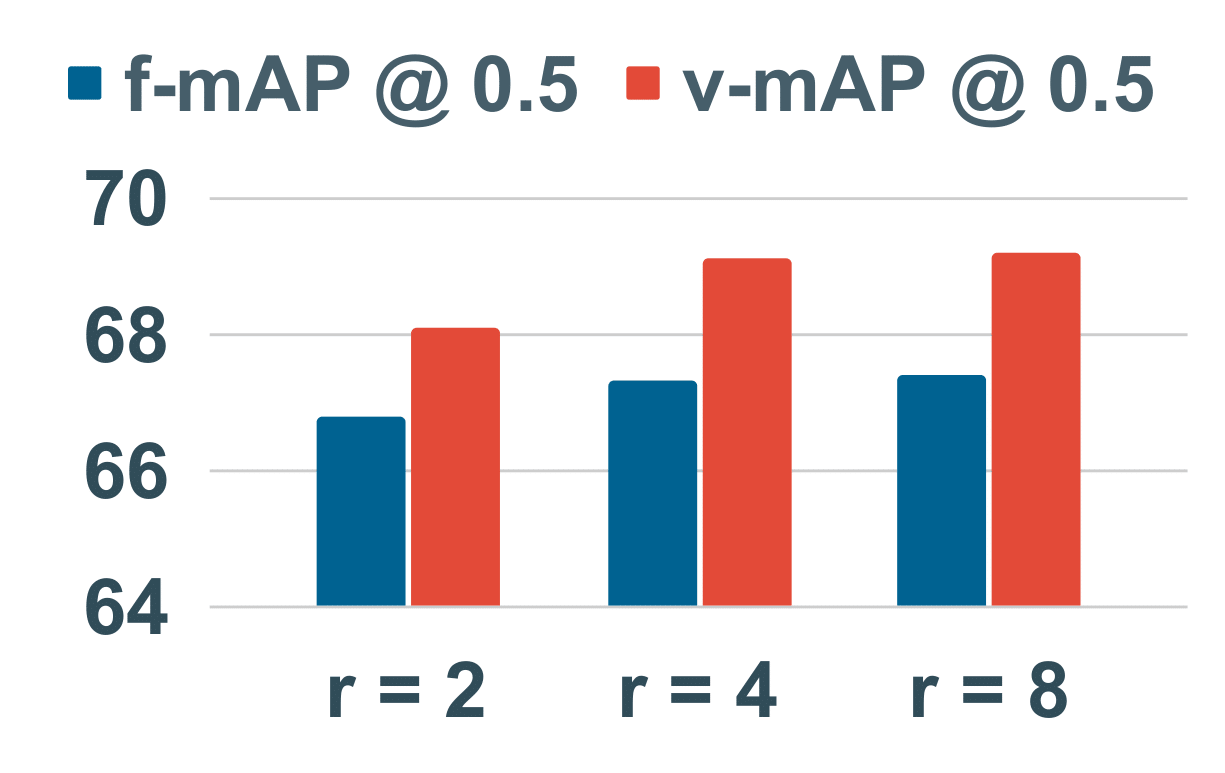}
    \caption{\textit{FFT Analysis:} Left: FFT filter on teacher(\textbf{T})-student(\textbf{S}) for SSL training. Right: Effect of radius (\textbf{r}) on FFT filter. Both are on UCF101-24 for 10\% labeled data. 
    }
    \label{fig:fft_mean_radius_ucf101}
\end{figure}

\paragraph{FFT as a high pass filter}
For a given frame $f$ of a video, the model performance $M(f;\theta)$ can be categorized in four main types: High-Confidence and High-Consistency (HCF-HC), High-Confidence and Low-Consistency (HCF-LC), Low-Confidence and High-Consistency (LCF-HC) and Low-Confidence and Low-Consistency (LCF-LC). The simple consistency loss from Equation \ref{eq:consistency_loss} focuses on regions with lower consistency (either HCF-LC or LCF-LC). The model is not able to predict confidently for LCF-LC samples due to lack of similar supervised training samples. 

Ideally we do not want to give high weight for LCF-LC samples in unsupervised setting as model is not able to predict anything for such samples. In contrary, for unsupervised setting we would prefer to have more weights on HCF-LC samples as model is confident but inconsistent. Using a high pass filter enables this, as it gives higher weight for high confidence (HCF) regions and filters out low confidence (LCF)  regions. This is demonstrated in Figure \ref{fig:heatmap}, where the general consistency loss gives equal weight to large background region and small action region, making it hard for network to learn with noise augmentation. Our FFT based approach gives higher weights on action regions (specifically the edges) which the model can improve on more than background.


\paragraph{Radius for FFT filter}
Fft-attention relies on the radius of the filter which affects the value for each pixel from FFT. While small radius looks at local window, it has higher sensitivity based on local changes. Conversely, larger radius looks at larger neighborhood but dilates the effect in return. We analyze different radius for FFT filter (Figure \ref{fig:fft_mean_radius_ucf101}) and found it to be robust within a range.



\section{Conclusion}
We present a unified semi-supervised active learning approach for spatio-temporal video action detection, particularly in scenarios where obtaining labels is costly. We show that using noise as augmentation to compute the informativeness of each sample improves the sample selection for active learning. We also introduce the use of FFT based high pass filter to focus more on relevant activity regions for SSL consistency. 
Our proposed approach is characterized by its simplicity and can be easily generalized to other dense prediction tasks in videos. 


\section{Acknowledgments}
This research is based upon work supported in part by the Office of the Director of National Intelligence (Intelligence Advanced Research Projects Activity) via 2022-21102100001 and in part by University of Central Florida seed funding. The views and conclusions contained herein are those of the authors and should not be interpreted as necessarily representing the official policies, either expressed or implied, of ODNI, IARPA, or the US Government. The US Government is authorized to reproduce and distribute reprints for governmental purposes notwithstanding any copyright annotation therein.

\small{
\bibliography{aaai24}
}
\end{document}